\documentclass{article}

\usepackage[dblblindworkshop, final]{neurips_2025}

\usepackage[utf8]{inputenc} 
\usepackage[T1]{fontenc}    
\usepackage{hyperref}       
\usepackage{url}            
\usepackage{booktabs}       
\usepackage{amsfonts}       
\usepackage{nicefrac}       
\usepackage{microtype}      
\usepackage{xcolor}         
\usepackage{graphicx}

\title{Photorealistic Inpainting for Perturbation-based Explanations in Ecological Monitoring}

\author{%
  Günel Aghakishiyeva \\
  Duke University\\
  Durham, NC \\
  \And
  Jiayi Zhou \\
  Duke University \\
  Durham, NC \\
  \And
  Saagar Arya \\
 Duke University \\
 Durham, NC \\
  \And
  Julian Dale \\
 Duke University \\
 Durham, NC \\
  \And
 James David Poling \\
 University of Agder  \\
 Grimstad, Norway \\
   \And
  Holly R. Houliston \\
 University of Cambridge  \\
  Cambridge, England \\
  \And
  Jamie N. Womble \\
  U.S. National Park Service  \\
  Department of Interior \\
\And
Gregory D. Larsen \\
  Alaska Spatial Science  \\
  Fairbanks, AK \\
\And
  David W. Johnston \\
 Duke University  \\
 Durham, NC \\
  \And
  Brinnae Bent \\
  Duke University \\
  Durham, NC \\
  \texttt{brinnae.bent@duke.edu} \\
}

\begin{document}

\maketitle

\begin{abstract}
Ecological monitoring is increasingly automated by vision models, yet opaque predictions limit trust and field adoption. We present an inpainting-guided, perturbation-based explanation technique that produces photorealistic, mask-localized edits that preserve scene context. Unlike masking or blurring, these edits stay in-distribution and reveal which fine-grained morphological cues drive predictions in tasks such as species recognition and trait attribution. We demonstrate the approach on a YOLOv9 detector fine-tuned for harbor seal detection in Glacier Bay drone imagery, using Segment-Anything-Model-refined masks to support two interventions: (i) object removal/replacement (e.g., replacing seals with plausible ice/water or boats) and (ii) background replacement with original animals composited onto new scenes. Explanations are assessed by re-scoring perturbed images (flip rate, confidence drop) and by expert review for ecological plausibility and interpretability. The resulting explanations localize diagnostic structures, avoid deletion artifacts common to traditional perturbations, and yield domain-relevant insights that support expert validation and more trustworthy deployment of AI in ecology.
\end{abstract}

\section{Introduction}
Despite growing interest in explainable AI (XAI), its use in ecological applications remains limited. Post-hoc methods developed by the machine learning community exist, and recent ecology reviews note both regulatory momentum toward explainability and XAI’s potential to improve ecological AI systems \citep{gevaert2022explainable,buchelt2024exploring}. Yet most deployed models in ecology remain “black boxes,” undermining transparency, reproducibility, and stakeholder confidence.

Ecological monitoring requires explainability techniques that provide actionable, domain-relevant insights. It is important to choose explainability methods that align well with human expert intuition to be paired with expert human review. Perturbation-based explanations are especially well-suited for ecological monitoring because they offer intuitive, human-understandable insights into model decisions. This is particularly valuable in conservation and land management settings, where experts need to understand not just what a model predicts, but why. Perturbation-based explanations can highlight subtle visual or environmental cues that influence classification, enabling better trust, error analysis, and actionable feedback in real-world monitoring scenarios. 

Perturbations are traditionally applied to an image by blurring or blacking out image regions or altering pixels to flip a model’s prediction \citep{fong2017interpretable}. A challenge with this approach is that they often generate inputs that are not ecologically meaningful or realistic. This presents a challenge in ecological monitoring, where explanations must preserve the semantic integrity of the scene and reflect plausible environmental variations. For example, obscuring part of an animal or habitat may technically alter a model's prediction, but it does not offer actionable insight for ecologists seeking to understand how an event like species co-occurrence contributes to classification or detection. For wider adoption of efficient AI workflows in the ecological community, explanations must be designed to reflect real-world ecological variability so that human experts can interpret and trust the explanations. In this study, we seek to use inpainting to generate realistic perturbation-based explanations for ecological monitoring.

\section{Related Work}

\subsection{Conventional Approaches to Perturbation-based Explanations}
Traditional perturbation approaches probe model behavior by masking, occluding, or blurring image regions. Occlusion sensitivity maps highlight pixels whose removal alters predictions \citep{zhou2015object}. Other common strategies, such as blurring and rule-based edits test reliance on texture, shape or contextual cues \citep{vermeire2022evidence}. Later work formally defined “evidence counterfactuals” as the minimal regions whose removal flips a label \citep{vermeire2022evidence}. Although simple and widely used, these perturbations tend to produce unrealistic, out-of-distribution (statistically different from natural examples in the dataset) images that limit their value in domains where plausibility is critical.

\subsection{Inpainting in Explainability}
Generative inpainting replaces masked regions with plausible content, producing perturbations that remain in-distribution. Medical imaging studies use inpainting to remove tumors with only image-level labels \citep{shvetsov2024coin}, replace lesions with realistic healthy tissue \citep{afshar2024ibo}, and fill object parts to yield more consistent saliency evaluations \citep{badisa2024ing}. More recent methods extend this idea with stratified inpainting metrics \citep{cohen2025meaningful}, region-constrained counterfactuals \citep{sobieski2024rvce}, and diffusion-based text-to-image editing \citep{jeanneret2023time}. These works demonstrate that inpainting produces more natural, semantically faithful explanations than naive masking, though applications to ecological monitoring remain unexplored.

\section{Methods}
\subsection{General Pipeline}
We developed a pipeline for inpainting-guided, perturbation-based explanations using two techniques: 1) object removal or replacement, and 2) background replacement. 

We demonstrate this pipeline on a YOLOv9 \citep{wang2024yolov9} model fine-tuned for pinniped (harbor seal, Phoca vitulina richardii) detection (modeling details provided in Appendix B). YOLOv9 was fine-tuned on a dataset of 1,897 aerial images taken via drone in Glacier Bay National Park and Preserve (data collection and labeling methods in Appendix A). 

Inference was conducted on our test set and detected bounding boxes were refined into binary segmentation masks using the Segment Anything Model (SAM, ViT-b checkpoint) \citep{kirillov2023segany}. At inference, detections were accepted at a confidence threshold of 0.40, and union masks were generated for all boxes above this threshold. These masks defined the regions to be removed or altered. Of the 76 test images, 44 (58\%) met these criteria and were used for inpainting.

For both object removal/replacement and background replacement, we evaluated four inpainting models on perceptual realism and semantic consistency to determine which to use for the pipeline (methods and results for analysis in Appendix D, all natural language prompts used for analysis in Appendix C). Based on visual review, Stable Diffusion yielded the most reliable and semantically plausible outputs; with 100 inference steps, a guidance scale of 20.0, and the DPMSolverMultistep scheduler, it consistently produced stable results. All subsequent experiments were therefore conducted with Stable Diffusion. The inpainting calls were executed via the Replicate API. To ensure reproducibility, we fixed the random seed at 42 for all experiments.

\subsection{Object Removal and Replacement}
YOLO detections refined with SAM were used to generate masks for seal regions. These regions were edited with generative inpainting models under two conditions: (1) removal, where seals were replaced with plausible background content such as ice or water, and (2) replacement, where seals were substituted with boats. 

\subsection{Background Replacement}

YOLO detections refined with SAM were used to produce seal masks, which were then padded  (3 pixels) and feathered (Gaussian blur radius 1) to ensure clean boundaries. New backgrounds were generated with generative inpainting models applied to blank canvases. The original seals were composited back onto these generated backgrounds.

\subsection{Evaluation}

We evaluated inpainting results both quantitatively and qualitatively. Quantitatively, we re-applied the seal detector to inpainted images and measured changes in prediction validity and confidence (Appendix E). Qualitatively, results were further evaluated by side-by-side expert inspection of original and perturbed images. Experts judged whether inpainted regions maintained ecological plausibility and whether edits provided interpretable explanations of the detector's behavior.

\section{Results}

\subsection{Seal Removal}
In a representative case (Figure 1-A), removing a seal and filling the region with plausible ice eliminated detections entirely, showing that the detector relied on the seal morphology rather than contextual background. Across 44 threshold-passing originals, 36 removals (81.8\%) resulted in no detections, giving a flip rate of 0.82. For the remaining 8 images, detections persisted with mean confidence 0.108 ± 0.239, down from 0.696 before perturbation, yielding an average confidence drop of 0.588 ± 0.251.
Qualitatively, removals often produced background fills that preserved the scene while eliminating seals. In some cases, removing one seal increased the detection confidence for another in the same image, showing context-dependent effects (Figure 1-B). These results confirm that the detector relies on intrinsic seal morphology; once contours were removed, predictions collapsed in nearly all cases.

\begin{figure}[htbp]
  \centering
  \includegraphics[height=.4\textheight,keepaspectratio]{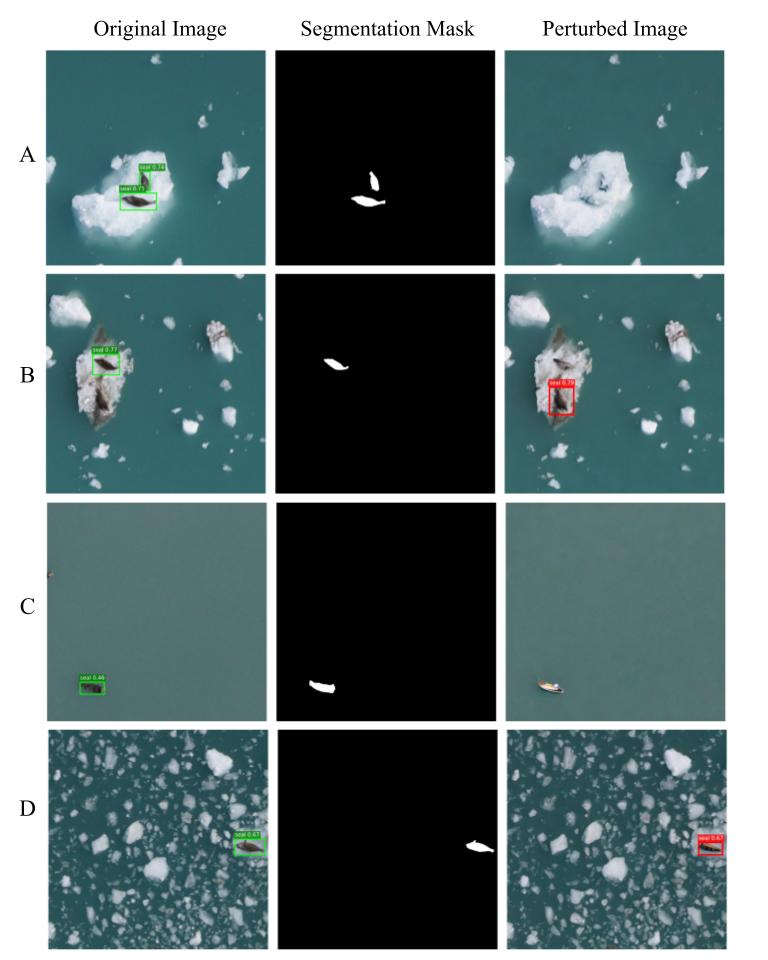}
  \caption{Inpainting-based perturbations for harbor seal detections. (A–D) Examples from the test set. Column 1 shows the original image with YOLOv9 detections, column 2 the SAM mask, and column 3 the perturbed image. (A) Removing a seal and filling with ice eliminates all detections. (B) Removing one seal lowers its detection but raises confidence for a neighbor. (C) Replacing a seal with a boat removes the detection. (D) A boat replacement is misclassified as a seal.}
  \label{fig:inpainting}
\end{figure}

\subsection{Seal Replacement}
In a representative case (Figure 1-C), replacing a seal with a boat removed the original detection, but in many cases the inpainting did not successfully insert a convincing boat. Across 44 threshold-passing originals, 20 produced plausible boat replacements. Of these, half (10) were not detected as seals, while the other half (10) were still misclassified as seals. The remaining 24 edits removed the seal without inserting a boat at all. Quantitatively, replacement produced flips in 26 cases overall (59.1\%). Flip rates were calculated with respect to all 44 attempted edits, including those where the replacement was visually implausible. When detections persisted, confidences averaged 0.245 ± 0.311 compared to 0.696 before perturbation, for an average confidence drop of 0.451 ± 0.315.

Qualitatively, even when boats were plausibly generated, half were still classified as seals (Figure 1-D), suggesting that the detector conflates certain textures and contours of boats with seal morphology. 
In some cases (example provided in Appendix F), visually similar replacements yielded opposite outcomes: one version produced no detection, while another sustained a 0.45 confidence. This sensitivity to subtle pixel-level differences highlights both the limitations of the generative model and the fragility of the detector when confronted with semantically novel objects.

\subsection{Background Replacement}
Background replacement tested robustness to context shifts by compositing seals onto new environments. In a typical case (Figure 2A-B), seals remained detected at high confidence even when placed against incongruent backdrops such as cityscapes or offices. Across 660 composites (the 44 threshold-passing originals each placed into 15 environments), detections were suppressed in only 42 cases (6.4\%), with an average confidence drop of 0.031 ± 0.188.

Flip rates varied across environments. Suppression was most frequent (Figure 2C) in beach (31.8\%), rocky (18.2\%), and winter (15.9\%) contexts, where sand, rocks, or snow occasionally obscured features. In contrast, desert, office, city, and several other environments yielded virtually no flips. Many contexts (Figure 2D), including clouds and studio, produced spurious detections where background elements such as cloud and studio items were also labeled as seals.
Qualitatively, background edits showed that the detector is largely invariant to environmental changes: seals were usually detected regardless of setting. Failures clustered in textured contexts, where the detector occasionally suppressed true detections or misclassified non-seal elements.

\begin{figure}[htbp]
  \centering
  \includegraphics[height=.4\textheight,keepaspectratio]{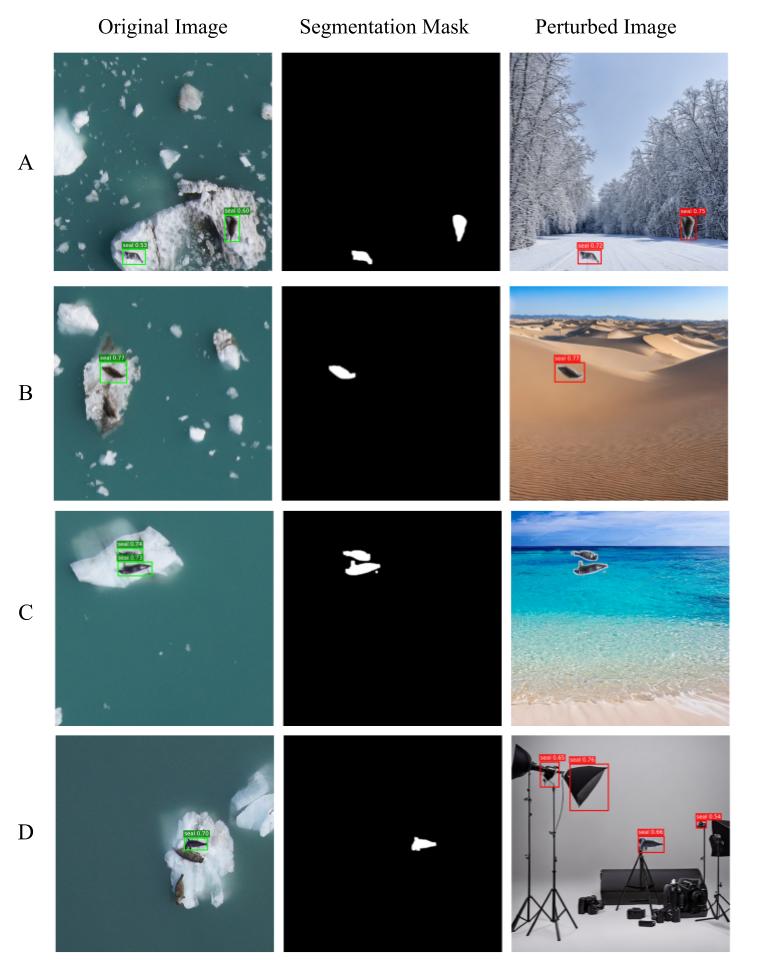}
  \caption{Background replacement for harbor seal detections. (A–D) Examples from the test set. Column 1 shows the original image with YOLOv9 detections, column 2 the SAM mask, and column 3 the perturbed image. (A–B) Seals remained detected at high confidence when placed into incongruent backgrounds such as winter landscapes and deserts. (C) Detection was occasionally suppressed in beach contexts. (D) Studio replacements produced spurious detections, with background objects misclassified as seals.}
  \label{fig:inpainting_background}
\end{figure}

\section{Discussion}
This study shows that inpainting-guided perturbations can produce ecologically meaningful “what-if” instances that improve our understanding of what the detector has actually learned. When we removed seals and plausibly filled the region, detections disappeared in approximately 82\% of cases and confidence dropped substantially, indicating that predictions hinge on fine-grained morphology rather than broad context. By contrast, compositing intact seals onto diverse backgrounds rarely suppressed detections overall (6.4\% flips), suggesting strong context invariance with pockets of brittleness in highly textured scenes (e.g., beach, rock, winter). Replacement edits revealed important confounds: boats occasionally remained (mis)classified as seals, revealing edge/texture overlaps that the detector treats as diagnostic. 

Compared with traditional perturbation techniques, photorealistic inpainting keeps images in-distribution, preserving ecological semantics so that both detectors and experts “see” plausible scenes. This yields explanations that are more actionable: removals test whether morphology is necessary, replacements probe “what else looks like the target?”, and background replacements assess dependence on context. Side-by-side expert review benefits from coherent scenes, facilitating judgments of plausibility and interpretability rather than reacting to obviously damaged inputs.

These findings translate into concrete practices for ecological monitoring. Removal edits can be used as a pre-deployment sanity check: detections should collapse when animal morphology is eliminated. Replacement outcomes identify hard negatives, such as the boats shown in our analysis, to prioritize targeted data curation and augmentation. Background replacements can be used to audit context robustness across different field conditions. Challenges identified here can help us consider actionable next steps for post-processing, data augmentation, and/or supplemental training data.

However, scaling this approach introduces challenges. The pipeline’s two subsequent steps, (1) mask refinement with SAM and (2) generative inpainting with Stable Diffusion, create computational overhead. The SAM mask prediction alone requires approximately 50 ms per prompt \citep{kirillov2023segany}, while Stable Diffusion inference can take ~5 days on a single A100 GPU to generate 50,000 samples \citep{rombach2022ldm}. This computational overhead suggests that inpainting-guided explanations should be used strategically rather than routinely during model validation phases, when investigating failure cases, or for generating training examples for human expert review. For large-scale workflows, we recommend using bounding boxes instead of segmentation (Appendix G). Future optimizations such as adopting faster segmentation models (FastSAM), more efficient inpainting architectures (SDXL-Turbo), and optimized inference parameters could reduce processing time by orders of magnitude, making routine deployment in large-scale monitoring programs more feasible.

Beyond computation, there are methodological caveats. Diffusion models carry implicit priors, and replacement edits can show instability, with similar objects producing inconsistent outcomes. Mask quality is critical: SAM overreach or underreach directly alters what is edited, and padding or feathering decisions affect boundaries. Results are parameter-sensitive (guidance scale, scheduler, seeds) and depend on detector calibration (confidence thresholds). Some backgrounds (e.g., office, city) are ecologically implausible; while useful for stress testing, their outcomes should be interpreted cautiously. Our scope, harbor seals in Glacier Bay, limits immediate generalization—replication across species, habitats, and seasons is needed. Future work could explore anatomy-targeted inpainting to isolate specific structures (e.g., head or neck) and map feature importance more precisely.

Through inpainting-based perturbations, removals confirmed reliance on shape, replacements revealed texture bias, and background replacements tested independence from scene context in our seal detection model. Photorealistic, ecologically coherent edits allow both automated systems and human experts to engage with plausible scenarios, fostering reliable interpretation, targeted curation, and more robust monitoring.

\bibliographystyle{plainnat}
\bibliography{neurips_2025}

\appendix

\section{Data Collection and Labeling Methods}
Drone surveys were conducted using a Wingtra One Gen II fixed-wing platform (Zurich, Switzerland) with vertical takeoff and landing (VTOL) capabilities imaging with a Sony Alpha 6100 APS-C camera (Tokyo, Japan) with a Sony E 20mm f/2.8 lens. Flight plans were created and carried out using WingtraPilot flight planning software. Flights operated at ~60–85 m altitude and ~9–22 m/s airspeed over regions that were historically sampled by occupied aircrafts. Drone operations were conducted under permit by NOAA and the NPS.

Flights in glacial fjords occurred along non-overlapping parallel transects oriented lengthwise through the glacial end of the fjord, with transects oriented approximately perpendicular to the glacier terminus. These flight plans were similar to those of historic surveys that used occupied aircrafts \citep{womble2020calibrating}, but were optimized to achieve high-density coverage of the inner regions of the fjords where seal densities are highest. Surveys from occupied aircrafts historically sampled the entire west arm of JHI along the same 12 transects year after year, maintaining a ~100-m buffer between photographs across transects and a ~20-m buffer between consecutive photographs along transects. Surveys from unoccupied aircrafts in 2023 and 2024 surveyed smaller gross extents with a series of nearly contiguous but not overlapping transects, maintaining a ~5-m buffer between photographs across transects and 65–70\% overlap between consecutive photographs along transects. Surveys from unoccupied aircrafts in glacial fjords consisted of 1–3 flights each using impromptu flight plans informed by the extent of floating ice habitat and drone performance in the prevailing weather conditions at the time of the survey. Surveys of terrestrial sites also consisted of parallel transects arranged in a high density to achieve a target of ~60\% overlap between photographs along transects and across transects.

 The training dataset was visually reviewed and manually thinned to remove photographs that did not include at least one positive instance of a harbor seal on floating ice. This was done to mitigate the risk of negative bias in model training, which can occur with an overabundance of negative training data. The resulting dataset images were each subdivided into tiles of 640×640 pixels. Each tile was manually inspected and annotated to mark all harbor seal locations using Labelme \citep{russell2008labelme}.

\section{YOLOv9 Model Setup}
Prior to training, images were pre-processed to ensure compatibility with the YOLOv9 pipeline. A grid search was conducted to identify optimal hyperparameters, resulting in the following configuration: model type = yolov9c.pt, epochs = 300, patience = 30, batch size = 16, image size = 640, optimizer = AdamW, learning rate = 0.0099, momentum = 0.9599, weight decay = 0.00047, rotation degrees = 0.95, mixup = 0.27, scale = 0.24, and translation = 0.055. On the held-out test set of 76 images containing 134 annotated instances, the model achieved a precision of 0.94 and a recall of 0.97. In terms of average precision, performance was very strong at a single threshold (mAP@50 = 0.98), while across stricter thresholds (mAP@50–95), the model achieved 0.67. Data from our held-out test set were used for explanations. 

\section{Natural Language Prompts}
We report below the exact natural-language prompts supplied to each model. Placeholders such as 〈class〉 were programmatically replaced by the detected object label (e.g. seal).
\subsection{Object Removal and Replacement}
Stable Diffusion Inpainting
Positive prompt (default):
“photorealistic natural background scene, seamlessly filled area, consistent lighting and perspective, no artificial boundaries or seams.”
Negative prompt (default):
“duplicate, distorted, glitch, blur, shadow, extra limbs, deformed, low quality, bad anatomy, seams, harsh edges, inconsistent lighting, artifacts, pixelated.”
FLUX.1-dev Inpainting
Positive prompt (default):
“natural coherent background environment, perfectly blended inpainting, no visible artifacts or object remnants, realistic lighting.”
SDXL Inpainting
Positive prompt (default):
“clean photorealistic background that flows naturally with the original scene context, seamless integration, professional quality.”
Negative prompt (default):
“duplicate, distorted, glitch, blur, shadow, extra limbs, deformed, low quality, bad anatomy, seams, harsh transitions, inconsistent texture, artifacts.”
Per-class negative prompt (auto-generated):
“〈class〉, duplicate, distortion, seams, artifacts, low quality.”

\subsection{Background Replacement}
When inpainting backgrounds around detected objects, the following descriptions were used verbatim:
forest: “dense green forest with tall trees, natural woodland environment.”
mountain: “majestic mountain landscape with rocky peaks and dramatic scenery.”
beach: “tropical beach with white sand and blue ocean waves.”
city: “modern urban cityscape with tall buildings and streets.”
desert: “vast sandy desert with rolling dunes under clear sky.”
countryside: “peaceful rural countryside with green fields and rolling hills.”
garden: “beautiful botanical garden with colorful flowers and lush plants.”
winter: “snowy winter landscape with snow-covered trees and ground.”
tropical: “tropical paradise with palm trees and exotic vegetation.”
rocky: “rugged rocky terrain with dramatic stone formations.”
sunset: “beautiful golden sunset sky with warm dramatic lighting.”
cloudy: “overcast sky with dramatic clouds and soft lighting.”
office: “modern office interior with clean professional environment.”
indoor: “clean indoor environment with neutral lighting.”
studio: “professional photography studio with neutral background.”
A.3 Background Generation for Compositing
For de novo background synthesis, the following default negative prompt was applied: 
“people, animals, text, logo, watermark, artifacts, distortion, low quality.”

\section{Comparison of Inpainting Models}
\subsection{Methods}
For both object removal/replacement and background replacement, we evaluated four inpainting models: LaMa \citep{suvorov2022lama}, FLUX.1 (dev checkpoint) \citep{blackforestlabs2025flux,labs2024flux}, Stable Diffusion Inpainting \citep{rombach2022ldm}, and SDXL \citep{podell2023sdxl}. Each model was tested across multiple parameter settings to assess stability and realism. The parameter ranges were as follows: LaMa with seed values \{42, 123\}; Stable Diffusion Inpainting with \texttt{guidance\_scale} $\in \{10, 15, 20\}$, \texttt{num\_inference\_steps} $\in \{50, 100, 250, 500\}$, and seed $\in \{42, 123\}$; FLUX.1-dev with \texttt{guidance\_scale} $\in \{10, 15, 20\}$, \texttt{strength} $\in \{0.5, 0.75, 1.0\}$, \texttt{num\_inference\_steps} $\in \{30, 40, 50\}$, and seed $\in \{42, 123\}$; and SDXL with \texttt{guidance\_scale} $\in \{25, 35, 50\}$, \texttt{num\_inference\_steps} $\in \{20, 40, 100, 250, 500\}$, \texttt{prompt\_strength} $\in \{0.5, 0.75, 1.0\}$, and seed $\in \{42, 123\}$. Masks and images were resized to model-specific defaults (Stable Diffusion $512\times512$, SDXL $1024\times1024$, FLUX $1024\times1024$, LaMa native). Prompts and negative prompts were designed to avoid unrealistic insertions and preserve ecological plausibility (prompts provided in Appendix~B). Unless otherwise stated, default schedulers were used (\texttt{DPMSolverMultistep} for Stable Diffusion, \texttt{K\_EULER} for SDXL, built-in flow scheduler for FLUX.1-dev). We evaluated models based on perceptual realism and semantic consistency to determine which to use for the pipeline.

\subsubsection{Results}
We tested four inpainting models: LaMa, FLUX.1-dev, SDXL, and Stable Diffusion. LaMa and FLUX.1-dev often produced unrealistic or incomplete fills, while SDXL generated photorealistic edits but with inconsistent quality. Based on visual review, Stable Diffusion provided the most reliable and semantically plausible outputs across parameter settings. Accordingly, Stable Diffusion was selected for all subsequent experiments (see Section~3.1 for details on final parameters).

\section{Evaluation}
Quantitatively, we re-applied the seal detector to inpainted images and measured changes in prediction validity and confidence. Prediction validity was assessed by the flip rate (FR), defined as the proportion of cases where no detections above the confidence threshold $\tau$ remained after perturbation:
\begin{equation}
\mathrm{FR}_{p} = \frac{1}{N} \sum_{i=1}^{N} \mathbf{1}\{\, n_i^{\prime}(p) = 0 \,\},
\end{equation}
where $n_i^{\prime}(p)$ is the number of detections in image $i$ after perturbation $p$ with confidence $\geq \tau$ (here $\tau = 0.40$).

We also measured the confidence drop (CD), computed using thresholded top confidences:
\begin{equation}
\mathrm{CD}_{p} = \frac{1}{N} \sum_{i=1}^{N} \big( s_i^{(\tau)} - s_i^{\prime (\tau)}(p) \big),
\end{equation}
where $s_i^{(\tau)} = s_i$ if $s_i \geq \tau$ and $0$ otherwise, and likewise for $s_i^{\prime (\tau)}(p)$.

\clearpage
\section{Pixel Level Sensitivity}
\begin{figure}[htbp]
  \centering
  \includegraphics[height=.3\textheight,keepaspectratio]{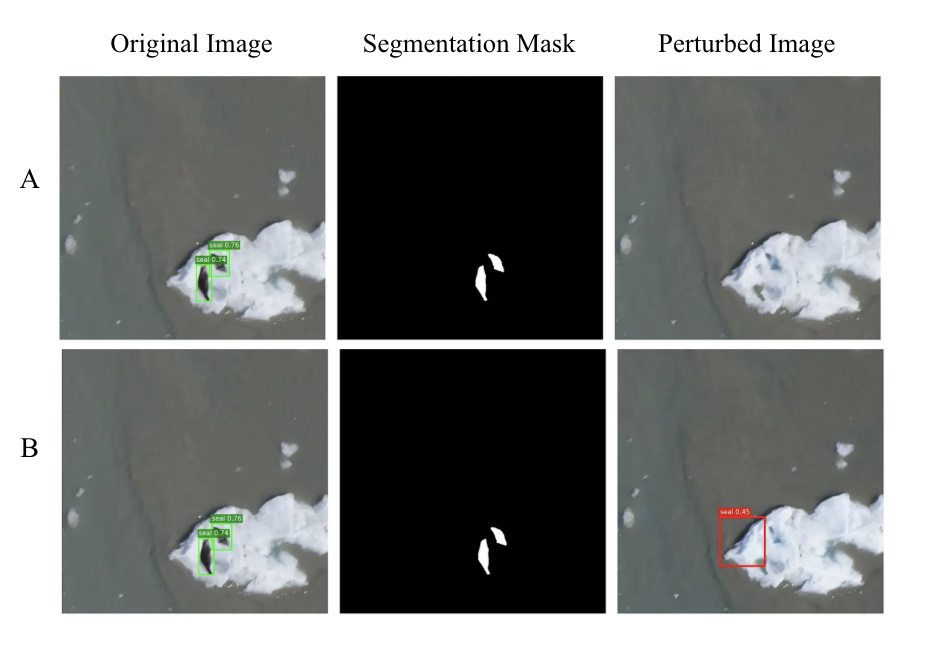}
  \caption{Pixel-level sensitivity in harbor seal detections. (A–B) Examples from the test set. Column 1 shows the original image with YOLOv9 detections, column 2 the SAM mask, and column 3 the perturbed image. (A) Removing seals eliminated detections. (B) A visually similar perturbation produced a spurious detection, showing detector sensitivity to pixel-level changes.}
  \label{fig:appendix}
\end{figure}

\clearpage
\section{Bounding Box vs. Segmentation Mask Runtime Comparison}

To address computational efficiency concerns, we compared our SAM-based segmentation approach with rectangular masks (bounding boxes) generated directly from YOLO detection outputs. We reran both removal and replacement experiments on the same set of 75 test images, keeping all other parameters constant.
During the removal phase, SAM required extra time to produce precise pixel-level masks, while bounding boxes were much faster to generate. For mask generation, bounding box masking reduced average removal runtime from 4.49s to 2.97s per image, yielding a 1.51× speedup. However, once masks were cached and the experiments were rerun for replacement edits, runtime differences disappeared, indicating similar timing during the inpainting step.

\begin{figure}[htbp]
  \centering
  \includegraphics[height=.3\textheight,keepaspectratio]{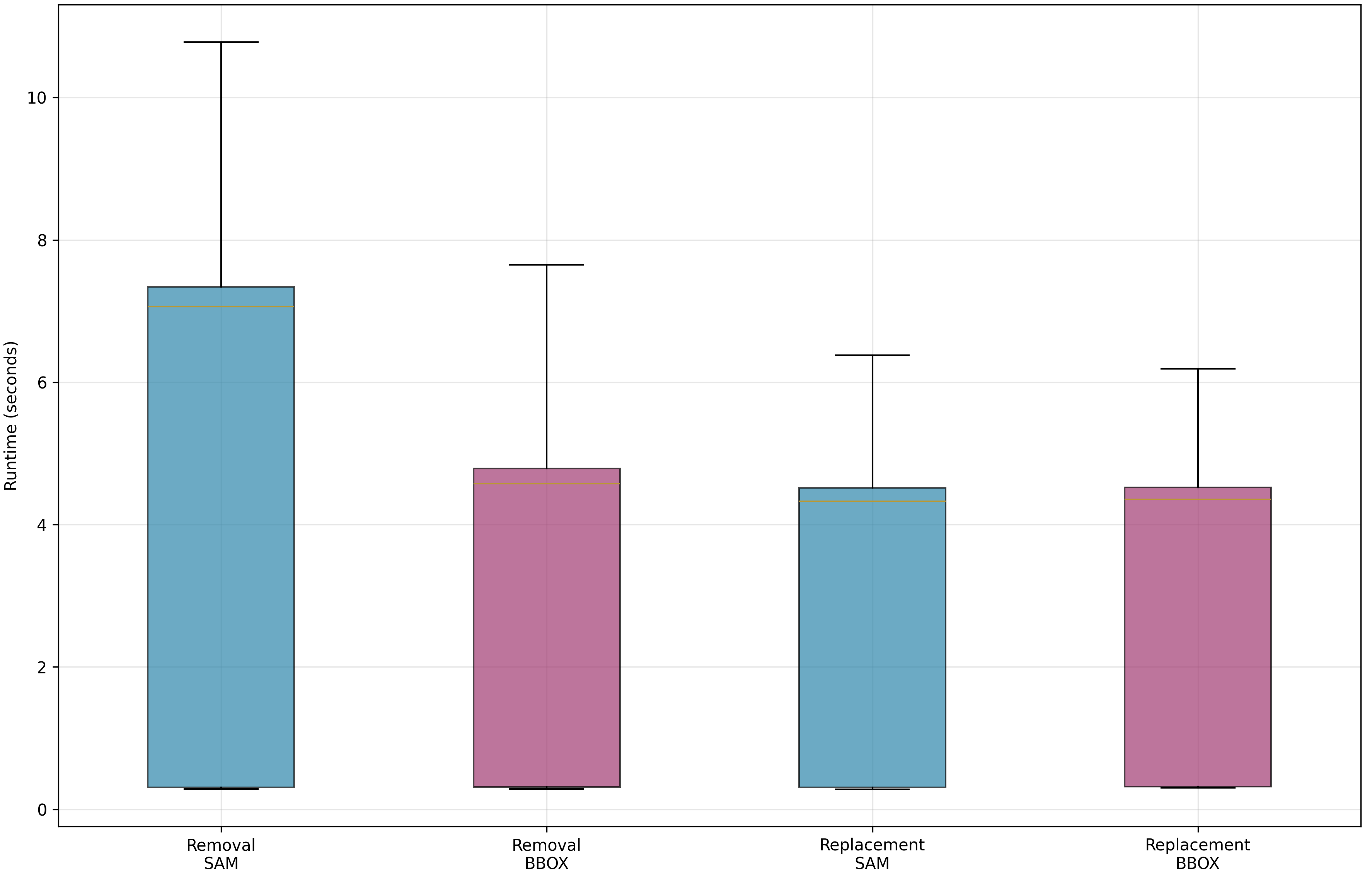}
  \caption{Runtime distribution for removal and replacement experiments using SAM-based segmentation and bounding box masking.}
  \label{fig:appendix}
\end{figure}

While both approaches perform similarly in inpainting once masks are generated, bounding boxes are better suited for large-scale workflows because they generate masks much faster at the start of the process.

\end{document}